\documentclass[sigconf]{acmart}
\usepackage{booktabs}
\usepackage{multirow}
\usepackage{xcolor}
\usepackage{listings}
\usepackage{listings}
\usepackage{graphicx}
\AtBeginDocument{%
  }

\setcopyright{acmlicensed}
\copyrightyear{2025}

\acmYear{2025}

\setcopyright{rightsretained}

\acmConference[KDD '25] {Proceedings of the 1st Workshop on "AI for Supply Chain: Today and Future" @ 31st ACM SIGKDD Conference on Knowledge Discovery and Data Mining V.2}{August 3, 2025}{Toronto, ON, Canada.}

\acmBooktitle{Proceedings of the 1st Workshop on "AI for Supply Chain: Today and Future" @ 31st ACM SIGKDD Conference on Knowledge Discovery and Data Mining V.2 (KDD '25), August 3, 2025, Toronto, ON, Canada}

\acmISBN{979-8-4007-1454-2/25/08}

\acmDOI{10.1145/XXXXXX.XXXXXX}

\acmISBN{978-1-4503-XXXX-X/2018/06}

\begin{document}

\title{C‑MAG: Cascade Multimodal Attributed Graphs for Supply Chain Link Prediction}

\author{Yunqing Li}
\authornote{Both authors contributed equally to this research.}
\orcid{1234-5678-9012}
\affiliation{%
  \institution{AAITC, Lenovo}
  \city{Morrisville}
  \state{North Carolina}
  \country{USA}}
\email{connie555514@gmail.com}

\author{Zixiang Tang}
\authornotemark[1]
\orcid{—} 
\affiliation{%
  \institution{AAITC, Lenovo}
  \city{Morrisville}
  \state{North Carolina}
  \country{USA}}
\email{zt2292@columbia.edu}

\author{Jiaying Zhuang}
\orcid{—} 
\affiliation{%
  \institution{AAITC, Lenovo}
  \city{Morrisville}
  \state{North Carolina}
  \country{USA}}
\email{jzhuang@lenovo.com}

\author{Zhenyu Yang}
\orcid{—} 
\affiliation{%
  \institution{AAITC, Lenovo}
  \city{Morrisville}
  \state{North Carolina}
  \country{USA}}
\email{zyang1@lenovo.com}

\author{Farhad Ameri}
\affiliation{%
  \institution{School of Manufacturing Systems and Networks, Arizona State University}
  \city{Mesa}
  \state{Arizona}
  \country{USA}}
\email{amerif@gmail.com}

\author{Jianbang Zhang}
\orcid{—} 
\affiliation{%
  \institution{AAITC, Lenovo}
  \city{Morrisville}
  \state{North Carolina}
  \country{USA}}
\email{zhangjb2@lenovo.com}

\renewcommand{\shortauthors}{Li et al.}


\begin{abstract}
Connecting an ever‐expanding catalogue of products with suitable manufacturers and suppliers is critical for resilient, efficient global supply chains, yet traditional methods struggle to capture complex capabilities, certifications, geographic constraints, and rich multimodal data of real‑world manufacturer profiles. To address these gaps, we introduce \textit{PMGraph}, a public benchmark of bipartite and heterogeneous multimodal supply‑chain graphs linking 8,888 manufacturers, over 70k products, more than 110k manufacturer–product edges, and over 29k product images. Building on this benchmark, we propose the Cascade Multimodal Attributed Graph (\textit{C‑MAG}), a two‑stage architecture that first aligns and aggregates textual and visual attributes into intermediate group embeddings, then propagates them through a manufacturer–product heterograph via multiscale message passing to enhance link prediction accuracy. \textit{C‑MAG} also provides practical guidelines for modality‑aware fusion, preserving predictive performance in noisy, real‑world settings.
\end{abstract}

\keywords{Graph Neural Network, Link Prediction, Mutlimodal Knowledge Graph, Graph Embedding, Manufacturer Relationship}


\maketitle

\section{Introduction}
Global supply chain disruptions\cite{panwar2022future} caused by pandemics, geopolitical tensions, and material shortages have underscored the importance of leveraging small and medium-sized enterprises (SMEs) and their potential capacities and capabilities more effectively. Beyond simply matching existing production capabilities, accurately predicting SMEs' potential to manufacture new products is essential for developing resilient and adaptable supply networks\cite{starly2020automating,li2021design}.

Three primary challenges complicate this objective. First, the absence of scalable, structured supply chain datasets limits the ability to model and analyze manufacturer-product relationships effectively. In most cases, publicly available data on manufacturing capabilities are sparse, heterogeneous, proprietary, and locked in unstructured formats \cite{jarvenpaa2019development,ismail2019manufacturing}. As a result, large-scale computational analysis becomes difficult. Without reliable data infrastructure, it is challenging to train AI models that can accurately match production requirements with suitable SME capabilities, let alone anticipate their ability to scale or diversify into new manufacturing domains.

The second major challenge lies in the complexity and rapidly evolving nature of manufacturing technologies. New processes, equipment, and materials are continuously introduced, making it difficult to maintain a current understanding of manufacturers’ capabilities \cite{teerndynamic,kamm2021knowledge}. Often, multiple methods exist for producing the same product, each requiring different technical resources and expertise. Effectively linking product requirements to manufacturers thus demands deep knowledge of process selection, material properties, and production technologies. \cite{wan2024making} This complexity presents a significant barrier to automating the identification of capable manufacturers, particularly among SMEs whose capabilities are rarely documented in structured, accessible formats.

Third, manufacturers in real‑world supply chain knowledge graphs (SC-KG) exhibit extensive heterogeneity—spanning geographic location, certifications, production capacity, and technical capabilities—which complicates the task of constructing unified representations and predicting both present and future product relationships~\cite{almahri2024enhancing,ameri2024open,10.1109/TPAMI.2018.2798607}. At the same time, fusing multi-modal data introduces its own set of challenges—misaligned embedding spaces, varying noise characteristics, incomplete modality coverage, and the risk of diluting key signals—demanding sophisticated alignment, weighting, and noise‑robust fusion strategies to maintain predictive performance \cite{liu2024alignrec}.

\begin{figure}[h]
  \centering
  \includegraphics[width=\linewidth]{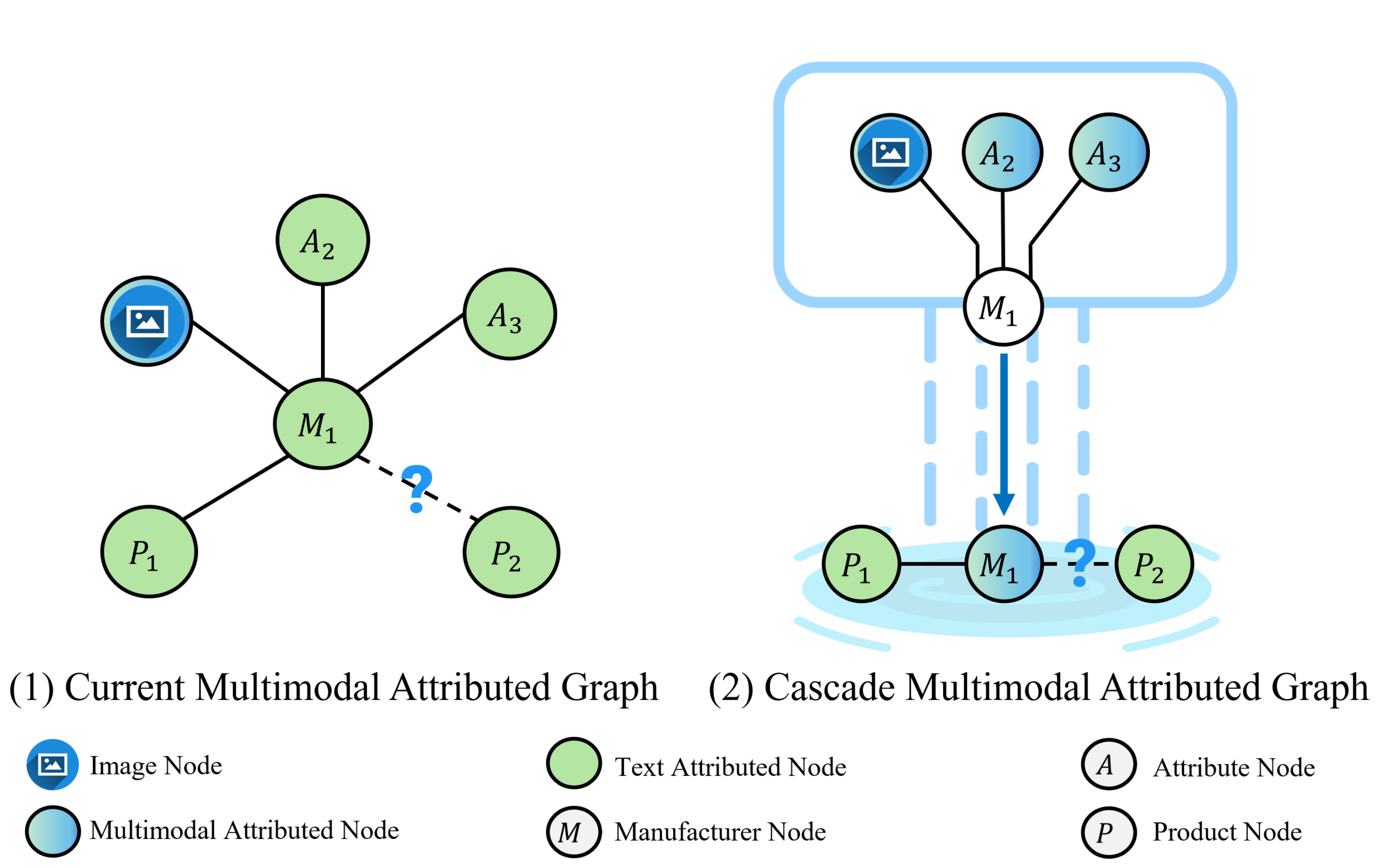}
  \caption{Overall architecture of the Cascade Multimodal Attributed Graph.}
  \label{fig:cmag_pipeline}
\end{figure}

To bridge these gaps, we propose three main contributions:

\begin{itemize}
  \item We release a suite of public‑use supply chain knowledge graphs \textit{PMGraph}\footnote{\textit{PMGraph} is publicly available at \url{https://huggingface.co/datasets/shawntzx/PMGraph}.} 
—including bipartite manufacturer–product graphs and heterogeneous multimodal knowledge graphs, and related variants—linking manufacturer metadata (capabilities, certifications, geographic locations) and visual assets to corresponding manufacturing products.
  
  \item We introduce Cascade Multimodal Attributed Graph (C-MAG) for manufacturer–product link prediction (Fig.~\ref{fig:cmag_pipeline}). In the top‐level graph, diverse manufacturer attribute nodes and image nodes—are algined and fused into a single group embedding. That embedding then flows into the lower heterograph, where it propagates between the manufacturer and product nodes via multiscale message passing, substantially boosting link‐prediction accuracy.

\item We present a graph‑based link‑prediction framework and comprehensive evaluation suite for quantifying the impact of visual features. Our benchmark reveals the trade‑offs of multimodal integration and delivers practical, modality‑aware fusion guidelines that safeguard predictive accuracy in noisy, real‑world supply‑chain scenarios.

\end{itemize}

For the rest of the paper, we review the related work in Sec.~\ref{sec:relate}. In Sec.~\ref{method}, the details of the proposed method are presented. In Sec.~\ref{sec:exp}, the experiments are conducted to demonstrate the effectiveness of our method. In Sec.~\ref{conclude}, the limitations and future work of our method are concluded.

\begin{figure*}[htbp]
  \centering
  \includegraphics[width=0.8\textwidth]{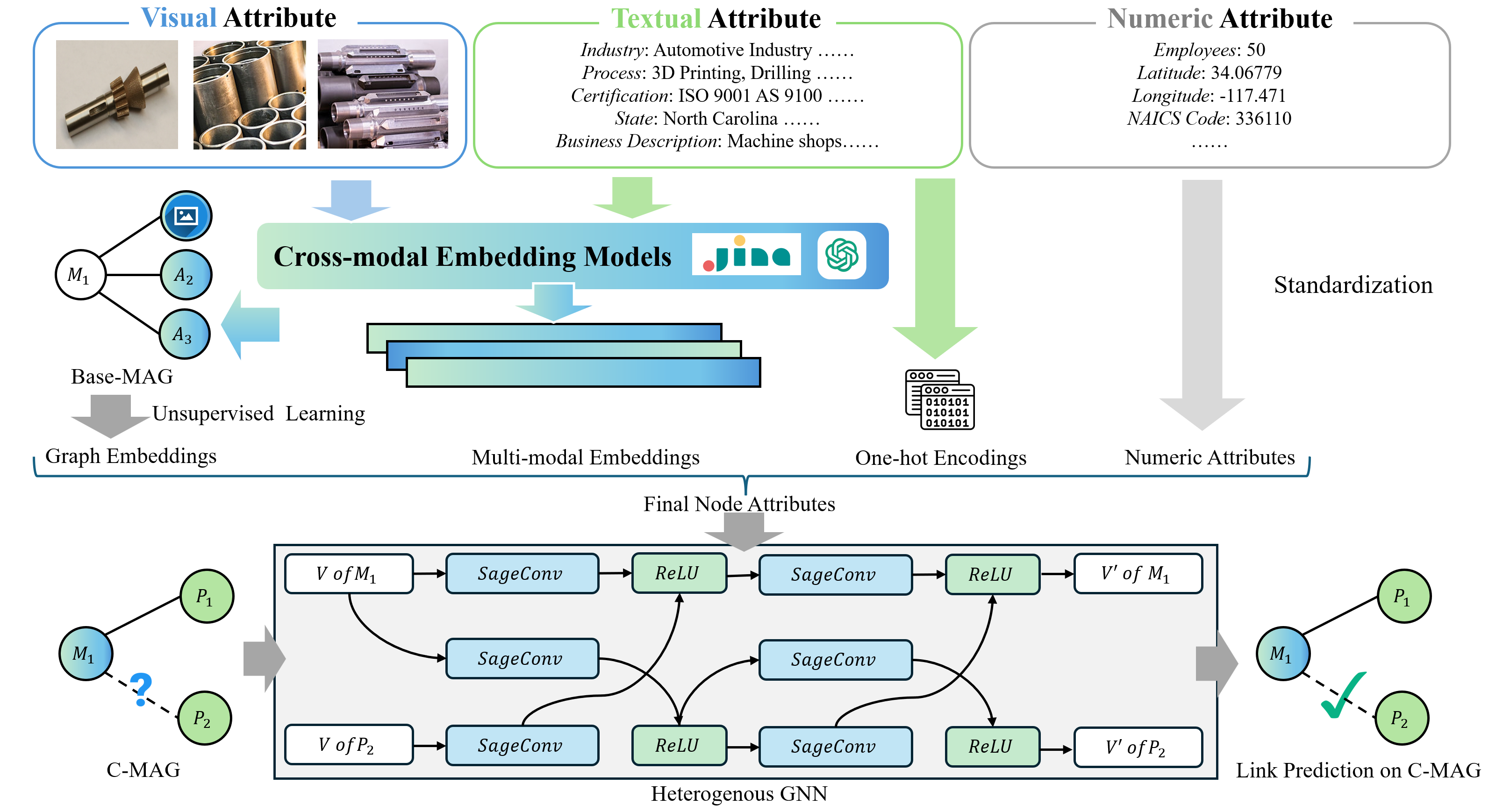}
  \caption{Overview of C-MAG pipeline}
  \label{fig:overview}
\end{figure*}

\section{Related Work}
\label{sec:relate}

\subsection{Graph Datasets}
\paragraph{Heterogeneous Graphs}

Heterogeneous graphs (or heterographs) extend standard graphs by allowing multiple node and edge types, enabling more faithful modeling of complex domains—such as supply chains—where entities and their relationships carry distinct semantics~\cite{dong2017metapath2vec, schlichtkrull2018rgcn}. Influential heterographs, like the Open Academic Graph (OAG), which merges 0.7 billion entities and 2 billion relations across papers, authors, venues, and institutions, and the diverse, large‑scale datasets of the Open Graph Benchmark (OGB) demonstrate the scale and domain heterogeneity that heterographs can capture~\cite{zhang2019oag,hu2020open}.

\paragraph{Multimodal Knowledge Graphs (MMKG)}
MMKGs extend heterogeneous graphs by enriching entities and relations with diverse data modalities—such as textual descriptions, images, and region‑level annotations—rather than relying solely on structured triples~\cite{zhu2022multimodal}. While heterographs model multiple node and edge types to capture domain heterogeneity, MMKGs embed unstructured or semi‑structured content directly into the graph, enabling joint reasoning over visual and textual signals~\cite{wang2022vqa}. For example, Visual Genome~\cite{krishna2017visual} builds scene‑level graphs from 100,000 images annotated with region descriptions, Richpedia~\cite{wang2020richpedia} links Wikidata entities to images via RDF‑based visual‑semantic relations, and MMEA‑UMVM~\cite{chen2023rethinking} provides 97 benchmark KGs with incomplete or ambiguous images to evaluate multimodal entity alignment.

\subsection{Attributed Graphs}
Attributed Graph (AG) is a directed multigraph where nodes and edges carry arbitrary key–value properties, enabling rich metadata annotation~\cite{pfeiffer2014attributed}. Unlike ontology‑bound RDF triples in knowledge graphs, AGs allow flexible property addition without a global schema. This flexibility supports fine‑grained entity–relation modeling and joint topology‑feature reasoning for analytics and machine learning.
~\cite{sheikh2019gat2vec}.

\paragraph{Text Attributed Graphs (TAG)}
Text‑attributed graphs enrich nodes (and optionally edges) with unstructured text. Key benchmarks include OGBN‑arxiv~\cite{hu2020open}, CS‑TAG’s eight large‑scale graphs across diverse domains~\cite{yan2023comprehensive}, and TEG‑DB’s unified text‑edge graph suite with a modular evaluation pipeline~\cite{li2024teg}.

\paragraph{Multimodal Attributed Graphs (MAG)}
A MAG enriches nodes and edges with heterogeneous feature modalities (e.g., structured data, text, images, audio). MAGB~\cite{yan2024graph} provides the first benchmark for MAG representation learning, and Kannan et al.~\cite{kannan2020multimodal} align scientific papers with code and images to illustrate text–visual fusion challenges. MMIEA~\cite{zhu2023mmiea} introduces a cross‑modal entity alignment model that addresses misalignment and missing‑modality issues in attributed graphs.

\subsection{Graph Based Link Prediction}

Link prediction (LP) in supply chain heterographs aims to infer missing or potential links by estimating the likelihood of connections between node pairs based on observed topology and attributes. Approaches divide into two paradigms:  Supervised methods train edge classifiers with cross‑entropy loss on known links using Graph Neural Network (GNNs) such as GCN~\cite{kipf2016semi}, GraphSAGE~\cite{hamilton2017inductive}, GAT~\cite{velickovic2017graph}, and HGT~\cite{hu2020heterogeneous}.  
Besides, unsupervised techiques, like GraphSAGE~\cite{hamilton2017inductive}, R‑GCN~\cite{schlichtkrull2018modeling}, and Node2Vec~\cite{grover2016node2vec},leverage contrastive objectives, negative sampling, or random‑walk embeddings to separate true from false edges without labels.

\subsection{Supply‑Chain Data Resources and GNN Applications}
Recent supply chain graph datasets include the SC-KG by Wichmann, Brintrup, and Baker et al., which models multi‑type relations (supplier–buyer, supplier–product, product–ingredient) across automotive and energy sectors for over 40,000 companies \cite{kosasih2024towards}; the Listed Companies Supply Chain Network of publicly traded Chinese firms capturing procurement and sales links for industry classification \cite{wu2023industry}; and SupplyGraph, a public FMCG production‑planning benchmark with temporal node features from a Bangladeshi company \cite{wasi2024supplygraph}. However, most of existing supply‑chain graphs are proprietary or limited to homogeneous relations within narrow domains or omit unstructured modalities (e.g., text, images)—underscoring the need for a publicly available, scalable, multimodal SC-KG.

Recent GNN‑based frameworks for manufacturer and supplier prediction include a Manufacturing Service Knowledge Graph that uses neighborhood aggregation and oversampling to handle imbalanced capability classification \cite{li2024manufacturing}; an ontological Knowledge Graph (KG) combined with Retrieval‑Augmented Generation for context‑aware supplier discovery \cite{10.1115/1.4067389}; hidden‑link prediction in automotive networks via GNNs augmented with Integrated Gradients for interpretability \cite{kosasih2022machine}; and a large‑scale automotive SC-KG leveraging structured and textual embeddings for alternative supplier recommendation \cite{tu2024using}. Despite their effectiveness, these frameworks largely rely on structural and textual embeddings, neglecting the rich visual assets and the alignment and noise challenges that arise when fusing heterogeneous modalities in supply‑chain graphs.

\section{Methodology}
\label{method}
To mitigate limitations of existing frameworks—which largely rely on structural and textual embeddings, neglecting rich visual assets and challenges of alignment and noise in multimodal SC-KGs, we propose C-MAG, a two-stage multimodal graph construction framework aimed at enhancing link prediction accuracy through hierarchical representation learning. The overall pipeline of our method is illustrated in Fig.~\ref{fig:overview}, highlighting the integration of visual, textual, categorical, and numeric attributes into node representations for LP.

\subsection{Stage 1: Base MAG and Auxiliary Pretraining}
\label{Stage1}

In the first stage, we construct a base-MAG capturing relationships among manufacturer nodes, attribute nodes (processes, certifications, materials, cities, states), and image nodes extracted directly from manufacturer websites.

\paragraph{Unified CLIP embeddings.} Manufacturer textual descriptions, attribute labels, and product images are encoded into a unified 768-D embedding space using Jina-CLIP-v1~\cite{koukounas2024jina}. Embeddings are then compressed to 32-D via Truncated Singular Value Decomposition (SVD)~\cite{abdi2007singular}, effectively reducing noise and computational complexity.

\paragraph{Unsupervised graph embedding.} Manufacturer embeddings are pretrained on this top level MAG via unsupervised link prediction. A two-layer GraphSAGE architecture (768 → 64 → 32) aggregates neighbor embeddings, applying linear transformations and ReLU activations. Positive edges (manufacturer → attribute/image) are contrasted against sampled negative edges using binary cross-entropy loss (BCE). An R-GCN variant is also validated to ensure architecture-agnostic embedding quality.

\subsection{Stage 2: Bipartite Manufacturer–Product Graph Design}
\label{Stage2}

In the second stage, we construct the bipartite manufacturer–product graph by initializing each node with the embeddings aggregated from the first‐stage attribute graph, thereby producing the final C‑MAG representation.

\paragraph{Textual and categorical embeddings.} Manufacturer and product textual attributes are processed with unified CLIP embeddings (768-D), compressed to 32-D with truncated SVD. Categorical metadata (business status, industry tags, certifications, NAICS codes) are one-hot encoded, numeric features (employee counts, geographic coordinates) are standardized, and the combined feature set is similarly compressed into 32-D embeddings via SVD.

\paragraph{Fusion and final multimodal graph assembly.} Manufacturer node embeddings from Stage 1 (32-D) are concatenated with Stage 2 textual, categorical, and numeric embeddings (64-D), forming enriched 96-D vectors. SVD further compresses these embeddings back to 64-D. Product nodes retain their original 64-D embeddings derived from textual and categorical data. The resulting a bipartite MAG encodes comprehensive manufacturer–product features, integrating structural, textual, categorical, numeric, and visual information for robust link prediction.

\subsection{Link Prediction}
\label{subsub:LP}
With these enriched node embeddings, we perform link prediction on manufacturer-product relationships using two heterogeneous Graph Neural Networks (heteroGNNs): HeteroSAGE and HeteroGAT. Initially, reverse edges (\textit{product $\rightarrow$ manufacturer}) are explicitly added to complement the forward edges (\textit{manufacturer $\rightarrow$ product}), thus creating a bidirectional heterogeneous graph structure that facilitates robust information propagation.

We implement two-layer architectures for both heteroGNN models, including intermediate nonlinear activations, dropout, and residual connections where applicable. Each model employs node-type-specific linear projection layers for embedding dimensionality alignment. To manage class imbalance, we utilize Weighted BCE ~\cite{mukhoti2020calibrating}.

\begin{table}
  \centering
  \caption{\textit{PMGraph} statistics.}
  \label{tab:pmgraph_stats}
  \begin{tabular}{llr}
    \toprule
    \textbf{Category} & \textbf{Type} & \textbf{\#} \\
    \midrule
    \multirow{4}{*}{Node types}
      & Manufacturer & 8,888 \\
      & Product      & 72,789 \\
      & Attribute    & 2,918 \\
      & Image        &  29,178 \\
    \midrule
    \multirow{3}{*}{Edge types}
      & manufacturer$\!\rightarrow$product    & 112,597 \\
      & manufacturer$\!\rightarrow$attribute  & 83,105  \\
      & manufacturer$\!\rightarrow$image      & 29,178  \\
    \bottomrule
  \end{tabular}
\end{table}

\section{Experiments and Analysis}
\label{sec:exp}
\subsection{Data Preparation}
\subsubsection{Data sources.} The raw graph corpus is assembled by merging (i) the SUDOKN SC-KG~\cite{Ameri2024OpenManufacturing} and (ii) the manufacturer directory and associated ontology of Li~\textit{etal.}\cite{LI2024100612}. Collectively, these sources comprise 8,888 unique manufacturer URLs (each offering multiple products) with extensive metadata, including employee counts, business descriptions, industry tags (up to 30 per manufacturer), process and material capabilities (up to 50 each), certifications, geographic coordinates, NAICS codes, and detailed product catalogs.

Product images are scraped from the manufacturer websites referenced in SUDOKN and then passed through a multimodal LLM–based quality filter to remove off‑topic or low‑resolution content (see Appendix~\ref{appendA}). After filtering, 145,888 images remain. Since including all image nodes can introduce noise, we conducted an ablation study to evaluate link‐prediction performance under different sampling ratios (See \ref{4.5.2}). In our main experiments, we randomly sample 20\% of the filtered images to enrich \textit{PMGraph}. The resulting graph statistics are presented in Table~\ref{tab:pmgraph_stats}.

\subsection{Graph Variants}
All six heterographs share the same manufacturer–product topology and compress node features to 64-D via truncated SVD. We evaluate two baseline models, \(\mathrm{AG}_{\mathrm{TFIDF}}\) and \(\mathrm{AG}_{\mathrm{JINA}}\), which use a flat bipartite graph with TF–IDF or Jina‑CLIP embeddings for textual features; two proposed cascade models, \(\mathrm{C\text{-}MAG}_{1}\) and \(\mathrm{C\text{-}MAG}_{2}\); and two ablation variants, \(\mathrm{FAG}_{1}\) (the flat counterpart of \(\mathrm{MAG}_{1}\)) and \(\mathrm{FMAG}_{2}\) (the flat counterpart of \(\mathrm{MAG}_{2}\)). This setup allows us to compare embedding methods, assess the benefit of a cascade architecture by contrasting \(\mathrm{FAG}_{1}\) versus \(\mathrm{C\text{-}MAG}_{1}\) and \(\mathrm{FMAG}_{2}\) versus \(\mathrm{C\text{-}MAG}_{2}\), and evaluate the impact of visual features by comparing \(\mathrm{C\text{-}MAG}_{2}\) versus \(\mathrm{C\text{-}MAG}_{1}\) and \(\mathrm{FMAG}_{2}\) versus \(\mathrm{FAG}_{1}\). The six variants are summarized in Table~\ref{tab:graph_variants}.

\begin{table}[t]
  \centering
  \caption{Variants of heterographs used in our experiments.}
  \label{tab:graph_variants}
  \begin{tabular}{@{}lccccc@{}}
    \toprule
    Variant                         & Hierarchy & Text Embedding & Images & Bipartite \\
    \midrule
    \(\mathrm{AG}_{\mathrm{TFIDF}}\)   & Flat      & TF–IDF         & No     & Yes       \\
    \(\mathrm{AG}_{\mathrm{JINA}}\)    & Flat      & Jina–CLIP      & No     & Yes       \\
    \(\mathrm{FAG}_{1}\)               & Flat      & Jina–CLIP      & No     & No        \\
    \(\mathrm{FMAG}_{2}\)              & Flat      & Jina–CLIP      & Yes    & No        \\
    \(\mathbf{C\text{-}MAG}_{1}\)           & 2‑layer   & Jina–CLIP      & No     & Yes       \\
    \(\mathbf{C\text{-}MAG}_{2}\)           & 2‑layer   & Jina–CLIP      & Yes    & Yes       \\
    \bottomrule
  \end{tabular}
\end{table}

\subsection{Experimental Setup}

\subsubsection{Unsupervised Pretraining}
\label{subsub:unsupervised}
Following the auxiliary pretraining in Sec.~\ref{Stage1}, we conduct random‐seed trials of a two‐layer GraphSAGE encoder (\(768\to64\to32\)) for up to 1000\,epochs with early stopping (patience \(=20\)). Optimization uses Adam (\(\mathrm{lr}=10^{-3}\), weight decay \(=10^{-5}\)), with batch sizes of 32 for attribute edges and 16 for image edges. To confirm architecture‐agnosticism, we repeat this procedure using a two‐layer R‐GCN under identical settings.

\subsubsection{Supervised Link Prediction}
\label{subsub:supervised}

Following most of the experimental setup of Lv et al.\ \cite{lv2021we}, we fine‑tune two heteroGNN architectures—HeteroSAGE and HeteroGAT (4 attention heads)—on both the bipartite manufacturer–product graph and the full heterograph including attribute and image relations. Each model consists of two projection layers with 128 hidden units and applies a dropout rate of 0.5. We split edges into train/validation/test sets in an 80/10/10\% ratio and use a 1:1 negative‑sampling ratio. Training runs for up to 1,000 epochs with early stopping (patience = 20) monitored on the validation ROC–AUC. For hyperparameter tuning, we grid‑search the learning rate over \(\{1,5\}\times\{10^{-6},10^{-5},10^{-4},10^{-3},10^{-2}\}\) and optimize a weighted binary cross‑entropy loss, selecting the configuration that maximizes validation ROC–AUC.

\paragraph{Flat‐graph ablations.} While the bipartite link‐prediction setup uses only manufacturer–product edges, the flat‐graph ablations (FAG\(_1\), FMAG\(_2\)) trains on the full heterograph (manufacturers, products, attributes, images) but evaluates exclusively on the manufacturer–product relation. Concretely, we apply randomly link split to manufacturer-product edges only, then strip attribute and image edges from the validation and test graphs so that only bipartite links are scored.  

All experiments run on NVIDIA GPUs with fixed random seeds for reproducibility.

\subsubsection{Evaluation metrics}
We assess link‐prediction performance on held‐out product–manufacturer pairs using two threshold independent metrics:

\paragraph{ROC-AUC} ($\mathrm{AUC}_{\mathrm{ROC}}$)  
The probability that a randomly chosen true edge receives a higher dot‐product score than a randomly chosen non‐edge:
\[
\mathrm{AUC}_{\mathrm{ROC}}
= \Pr\bigl[s_{i_+} > s_{i_-}\bigr],
\]
where $s_i$ are model scores and $y_i\in\{0,1\}$ are binary labels for each test pair, and $i_+$ and $i_-$ index a random positive and negative example, respectively.

\paragraph{PR-AUC} ($\mathrm{AUC}_{\mathrm{PR}}$)  
The area under the precision–recall curve, summarizing the trade‐off between precision and recall across thresholds. Let
\[
\mathrm{Precision}(t) = \frac{\mathrm{TP}(t)}{\mathrm{TP}(t) + \mathrm{FP}(t)},\quad
\mathrm{Recall}(t) = \frac{\mathrm{TP}(t)}{\mathrm{TP}(t) + \mathrm{FN}(t)},
\]
then
\[
\mathrm{AUC}_{\mathrm{PR}}
= \int_{0}^{1} \mathrm{Precision}\bigl(\mathrm{Recall}^{-1}(r)\bigr)\,\mathrm{d}r.
\]

\begin{table}
  \centering
  \caption{ROC‑AUC and PR‑AUC (\%) for hetero‑GNNs across graph variants (with base-MAGs pre-trained on GraphSAGE). Best and second‑best scores in each column are highlighted in \textbf{bold} and \underline{underline}, respectively.}
  \label{tab:auc_scores_main}
  \begin{tabular}{@{}lcccc@{}}
    \toprule
    Variant & \multicolumn{2}{c}{HeteroSAGE} & \multicolumn{2}{c}{HeteroGAT} \\
            & ROC‑AUC & PR‑AUC       & ROC‑AUC       & PR‑AUC       \\
    \midrule
    \(\mathrm{AG}_{\mathrm{TFIDF}}\) & 54.60       & 56.38       & 70.92       & 70.11       \\
    \(\mathrm{AG}_{\mathrm{JINA}}\)  & 66.46       & 60.58       & 72.49       & 71.02       \\
    \(\mathrm{FAG}_{1}\)             & 63.63       & 58.20       & 72.49       & 70.94       \\
    \(\mathrm{FMAG}_{2}\)            & 57.11       & 51.90       & 71.55       & 69.63       \\
     \(\mathbf{C\text{-}MAG}_{1}\)             & \underline{66.88} & \underline{60.94} & \textbf{75.46} & \underline{73.78} \\
     \(\mathbf{C\text{-}MAG}_{2}\)             & \textbf{70.58} & \textbf{66.09} & \underline{73.49} & \textbf{74.30} \\
    \bottomrule
  \end{tabular}
\end{table}

\subsection{Link Prediction Performance}
\label{sec:LPP}

Table~\ref{tab:auc_scores_main} shows that the cascade variants $\mathrm{C\text{-}MAG}_1$ and $\mathrm{C\text{-}MAG}_2$ occupy the top two positions for both HeteroSAGE and HeteroGAT, clearly outperforming the flat $\mathrm{AG}_{\mathrm{JINA}}$ and $\mathrm{AG}_{\mathrm{TFIDF}}$ baselines. This demonstrates that enriching the bipartite manufacturer–product graph with manufacturer–attribute relations via a cascade architecture can improve link‐prediction, and that Jina‑CLIP embeddings capture manufacturing semantics more effectively than TF–IDF. 

C‑MAGs also outperform their flat counterparts (\(\mathrm{FAG}_1\) and \(\mathrm{FMAG}_2\)) due to their hierarchical design: they first aggregate attribute information and manufacturer–attribute relations into the manufacturer nodes via an unsupervised link‑prediction step, and then propagate these enriched partial embeddings into the manufacturer–product prediction network. This two‑stage integration captures both attribute and topological context more effectively than the flat models, which fuse all information in a single step.

We observe that adding product image nodes does not consistently improve LP: the flat image‐augmented model (\(\mathrm{FMAG}_2\)) underperforms its non‑visual counterpart (\(\mathrm{FAG}_1\)), and the cascade variant \(\mathrm{C\text{-}MAG}_2\) does not always outperform \(\mathrm{C\text{-}MAG}_1\). This likely stems from residual noise in the filtered image set—despite removing irrelevant or low‑quality images, spurious visual features can still degrade the learned embeddings. Nevertheless, the hierarchical, two‑stage design of \(\mathrm{C\text{-}MAG}_2\) helps mitigate this issue: by first refining manufacturer node representations with image information and then using those enriched embeddings for manufacturer–product prediction, it achieves greater robustness to noisy vision data than the flat \(\mathrm{FMAG}_2\) model.

\subsection{Ablation Studies}
\subsubsection{Ablation of Auxiliary Pretraining Encoder}
\label{4.5.1}
To assess whether our gains derive from the GraphSAGE architecture or the graph design itself, we compare unsupervised pretraining with two‐layer GraphSAGE against R‐GCN on the MAG\(_2\) variant. Table~\ref{tab:mag_auc_scores} reports downstream ROC‑AUC and PR‑AUC for each hetero‑GNN.

\begin{table}
  \centering
  \caption{ROC‑AUC and PR‑AUC (\%) for hetero‑GNNs on \(\mathbf{C\text{-}MAG}_1\) and \(\mathbf{C\text{-}MAG}_2\) (with base-MAGs pre-trained on R-GCN). Best scores in each column are highlighted in \textbf{bold}.}
  \label{tab:mag_auc_scores}
  \begin{tabular}{@{}lcccc@{}}
    \toprule
    Variant & \multicolumn{2}{c}{HeteroSAGE} & \multicolumn{2}{c}{HeteroGAT} \\
            & ROC‑AUC & PR‑AUC       & ROC‑AUC       & PR‑AUC       \\
    \midrule
    \(\mathbf{C\text{-}MAG}_1\) & \textbf{70.41} & \textbf{65.90} & \textbf{74.84} & 72.59       \\
    \(\mathbf{C\text{-}MAG}_2\) & 67.97         & 62.15         & 74.42         & \textbf{72.70} \\
    \bottomrule
  \end{tabular}
\end{table}

With R‑GCN, the cascade models $\mathrm{C\text{-}MAG}_1$ and $\mathrm{C\text{-}MAG}_2$ still outperform all other variants; however, integrating multimodal features can sometimes degrade LP performance, likely due to residual noise in the filtered image data.

\subsubsection{Ablation of Images Sampling Ratio}
\label{4.5.2}
\begin{figure}[h]
  \centering
  \includegraphics[width=\linewidth]{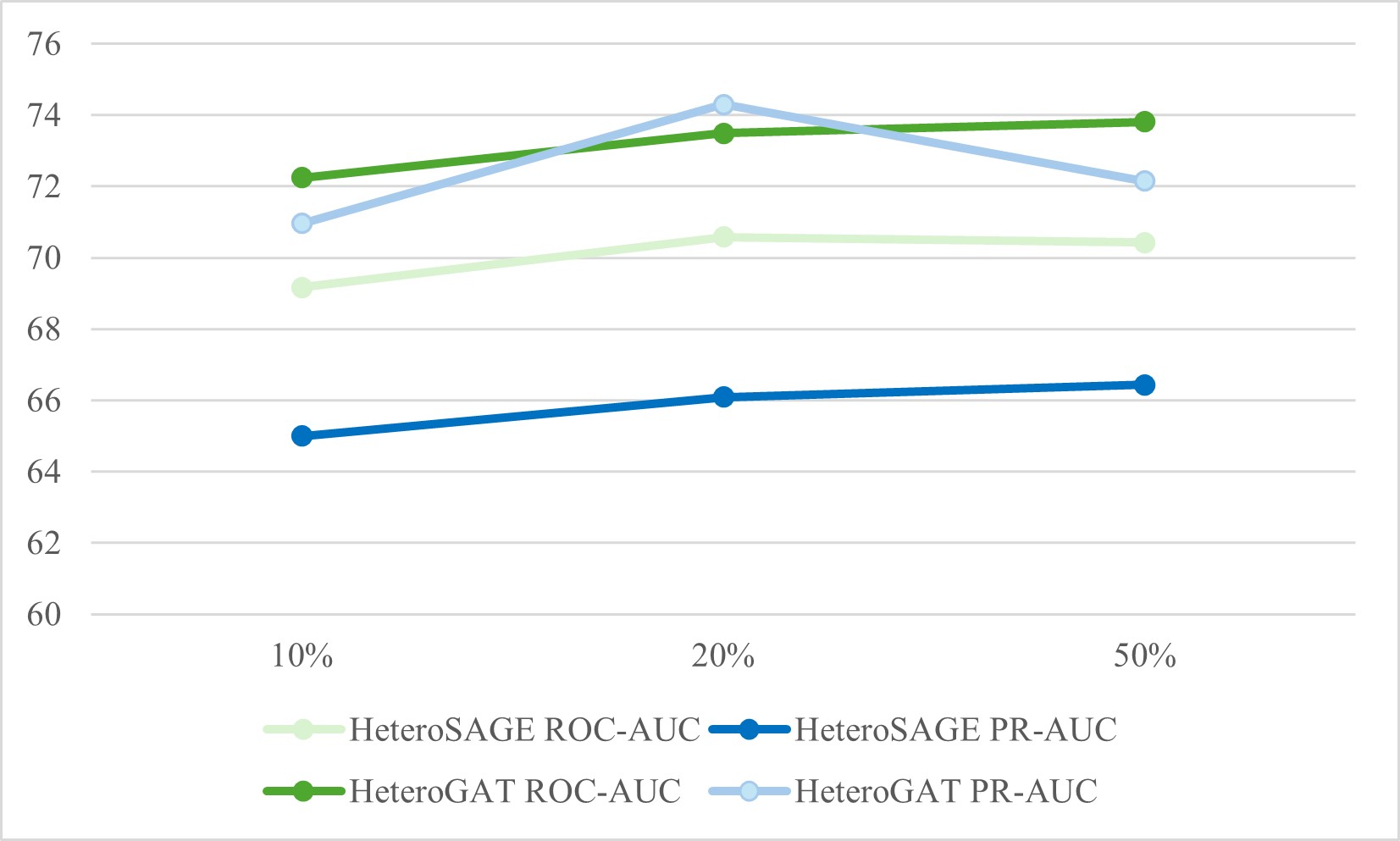}
  \caption{LP Performance with ROC‑AUC(\%) and PR‑AUC(\%) under different image sampling rates on for C‑MAG$_2$.}
  \label{fig:ablation_sampling}
\end{figure}

To evaluate how the proportion of image nodes affects link‐prediction in the cascade model, we ablate the image sampling ratio for C‑MAG$_2$ at 10\%, 20\%, and 50\%. Figure~\ref{fig:ablation_sampling} presents the ROC‑AUC and PR‑AUC trends obtained with the HeteroSAGE and HeteroGAT encoders.

Performance for both architectures peaks at 20\% sampling, with HeteroGAT consistently outperforming HeteroSAGE. Increasing sampling beyond 20\% yields only marginal improvements—likely a result of residual noise in the filtered images—but C‑MAG$_2$’s hierarchical cascade maintains strong LP accuracy, outperforming all flat variants and demonstrating robustness to noisy visual data.

\section{Conclusion and Future Work}
\label{conclude}

In this work, we introduce \textit{PMGraph}, a large‐scale, heterogeneous, multimodal supply‐chain benchmark linking 8,888 manufacturers, over 70,000 products, and over 29,000 product images, and propose C-MAG architecture customized for manufacturer–product LP. By enriching the bipartite manufacturer–product graph with manufacturer–attribute relations in a two‐stage cascade—first aggregating textual and visual attributes into intermediate embeddings, then propagating these through the full heterograph—C‑MAG achieves state‐of‐the‐art ROC‑AUC and PR‑AUC under both HeteroSAGE and HeteroGAT, substantially outperforming other graph variants. Our ablations show that C‑MAG’s hierarchical fusion is robust to noisy visual data, peaking at moderate image sampling ratios while maintaining strong performance even as image noise increases. These results demonstrate the importance of modality‐aware, staged integration of heterogeneous information for robust SC-KG modeling. Moreover, the C‑MAG framework readily generalizes beyond manufacturer–product matching to other supply‑chain inference tasks—such as component prediction, domain‑specific LP, and e-commerce recommendation system.

Future work will investigate adaptive image‐filtering strategies using advanced vision–language models (e.g., GPT‑o3~\cite{openai2025o3o4mini}, Claude Sonnet~\cite{anthropic2025claudesonnet4}) and explore dynamic cascade depths to better accommodate varying data quality \cite{tian2023asa,gao2023dynamic}. We will also extend evaluations to a wider range of heterogeneous GNN architectures \cite{shi2022heterogeneous,fu2020magnn} and conduct systematic ablations over key hyperparameters to optimize model performance in diverse, real‐world supply‑chain scenarios.


\bibliographystyle{ACM-Reference-Format}
\bibliography{sample-base}

\appendix

\section{Visual Data Extraction}
\label{appendA}
\subsection{Image Collection}

We develop a scalable, high-throughput pipeline to systematically collect product images from a curated set of over 8,000 manufacturer websites. Each target URL is first normalized to ensure syntactic consistency before initiating HTTP requests to retrieve the corresponding webpage content. To robustly accommodate a wide range of webpage structures—including malformed markup or embedded binary artifacts—the retrieved HTML is processed using a fault-tolerant, two-stage parsing strategy.

Following content extraction, all intra-domain hyperlinks are identified and subjected to a lexical filtering mechanism. Specifically, links are retained if they contain keywords indicative of product-related content (e.g., ``product'',``item'' ``catalog'', ``gallery'', ``prod'') and discarded if they correspond to non-product sections (e.g., ``about'', ``contact'', ``blog'', ``news'', ``login'', ``signup''). In cases where few product-specific links are detected, a supplementary set of additional internal pages is also explored to improve recall.

The homepage and selected product-relevant pages are then scanned for image elements. All discovered image URLs are resolved to their absolute forms, downloaded as binary files, and stored in structured subdirectories organized by domain. Filenames are semantically generated based on associated textual metadata, such as alt attributes, or default to the original filenames when descriptive information is unavailable—ensuring contextual relevance and traceability.

By integrating structured domain ingestion, robust HTML parsing, keyword-driven link prioritization, and hierarchical image storage, this pipeline enables efficient, scalable, and high-precision harvesting of product imagery across a broad landscape of industrial web sources.

\subsection{Image Filtering}

\paragraph{Vision–Language Filtering}
Subsequently, a vision–language filtering step is introduced using the \texttt{Gemini-2.0-Flash} model. For each image in a manufacturer’s directory, we issue the following templated prompt (with \texttt{product\_list\_str} dynamically populated from our dataset):

\begin{itemize}
  \item You are an image verification assistant. Given an image and a list of products (\texttt{product\_list\_str}), respond with exactly one word: “Yes” or “No.”
  \item Reply “Yes” only if you are absolutely certain the image shows one of the listed products.
  \item Reply “No” if you are uncertain, if the product is not in the list, or if image quality is too low to decide.
\end{itemize}

\end{document}